\documentclass[10pt,twocolumn,letterpaper]{article}

\usepackage[pagenumbers]{cvpr} %

\usepackage{xspace}
\newcommand{\shortname}{NaTex\xspace}

\usepackage{blindtext}
\usepackage{graphicx}
\usepackage{multirow}
\usepackage{bbding}
\usepackage{pifont}

\definecolor{cvprblue}{rgb}{0.21,0.49,0.74}
\usepackage[pagebackref,breaklinks,colorlinks,allcolors=cvprblue]{hyperref}

\def\paperID{*} %
\def\confName{CVPR}
\def\confYear{2026}

\title{\shortname: Seamless Texture Generation as Latent Color Diffusion}
\author{
    Zeqiang Lai$^{1,2\star}$
    , Yunfei Zhao$^{2\star}$ 
    , Zibo Zhao$^{2}$ 
    , Xin Yang$^{2}$ \\
     Xin Huang$^{2}$ 
    , Jingwei Huang$^{2}$ 
    , Xiangyu Yue$^{1\ddagger}$ 
    , Chunchao Guo$^{2\ddagger}$ 
    \\ 
	$^1${MMLab, CUHK} \quad $^2${Tencent Hunyuan}\\
    \url{https://natex-ldm.github.io}
}

\begin{document}

\twocolumn[{%
\renewcommand\twocolumn[1][]{#1}%
\maketitle
\begin{center}
    \vspace{-6mm}
    \centering
    \captionsetup{type=figure}
    \includegraphics[width=\linewidth]{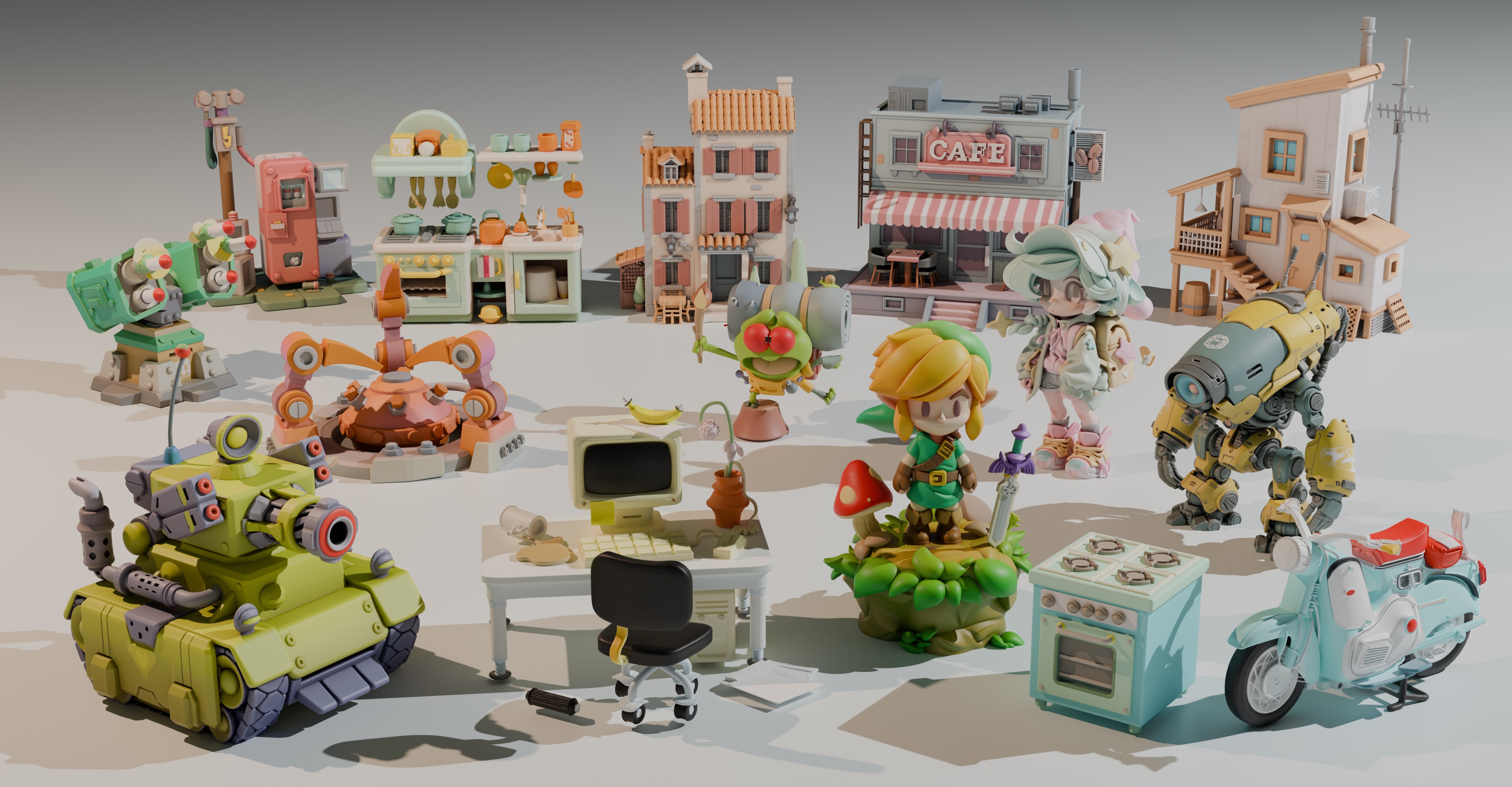}
    \captionof{figure}{
    High-quality textured 3D assets generated by \shortname from a single image (Geometry from Hunyuan3D 2.5~\cite{lai2025hunyuan3d}.)
    }
    \label{fig:teaser}
\end{center}%
}]

\def\thefootnote{}\footnotetext{$\star$ Equal contribution. $\ddagger$ Corresponding authors.} 

\begin{abstract}

We present \shortname, a native texture generation framework that predicts texture color directly in 3D space. In contrast to previous approaches that rely on baking 2D multi-view images synthesized by geometry-conditioned Multi-View Diffusion models (MVDs), \shortname avoids several inherent limitations of the MVD pipeline. These include difficulties in handling occluded regions that require inpainting, achieving precise mesh-texture alignment along boundaries, and maintaining cross-view consistency and coherence in both content and color intensity. 
\shortname features a novel paradigm that addresses the aforementioned issues by viewing texture as a dense color point cloud. 
Driven by this idea, we propose latent color diffusion, which comprises a geometry-aware color point cloud VAE and a multi-control diffusion transformer (DiT), entirely trained from scratch using 3D data, for texture reconstruction and generation. 
To enable precise alignment, we introduce native geometry control that conditions the DiT on direct 3D spatial information via positional embeddings and geometry latents.
We co-design the VAE–DiT architecture, where the geometry latents are extracted via a dedicated geometry branch tightly coupled with the color VAE, providing fine-grained surface guidance that maintains strong correspondence with the texture.
With these designs, \shortname demonstrates strong performance, significantly outperforming previous methods in texture coherence and alignment. Moreover, \shortname also exhibits strong generalization capabilities, either training-free or with simple tuning, for various downstream applications, \eg, material generation, texture refinement, and part segmentation and texturing. 

\end{abstract}

\section{Introduction}

The creation of realistic and diverse materials is a cornerstone of modern computer graphics, directly governing the visual fidelity of everything from cinematic visual effects to immersive virtual worlds. However, the manual creation of textures remains a profound bottleneck—an artisanal process that is both time-consuming and requires deep expertise. This challenge has catalyzed a paradigm shift from manual creation to automated generation, seeking to develop generative systems that can produce high-quality, physically-plausible texture efficiently and at scale.

To meet the high standards, multi-view texturing has become the de facto approach in numerous research studies~\cite{hunyuan3d2025hunyuan3d,zhao2025hunyuan3d,zhang2024clay,he2025materialmvpilluminationinvariantmaterialgeneration,huang2024mvadapter} as well as commercial products~\cite{tripo,rodin,meshy,hitem3d,lai2025hunyuan3d}. The concept is straightforward. It first generates multi-view images that align the input geometry from different viewpoints. Then, using the camera information from these viewpoints, a deterministic backprojection process is employed to reconstruct 3D textures from the 2D views.
One of the key advantages of this paradigm is that we can leverage pre-trained image generative models~\cite{rombach2022high, li2024hunyuandit, flux2024} and accompanying techniques~\cite{brooks2022instructpix2pix, ye2023ip-adapter, cao2023masactrl, zhang2023adding} to generate multi-view images, which form the foundation for the high quality and diversity of the generated textures.

\begin{figure}[t] 
  \centering
  \includegraphics[width=\linewidth]{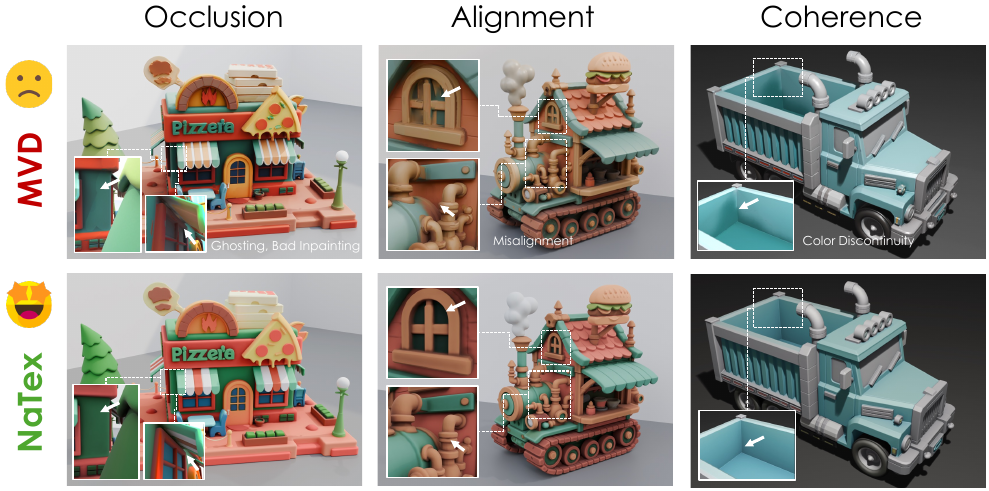}
  \caption{Illustration of the fundamental challenges in multi-view diffusion (MVD) texturing, compared with the proposed NaTex.} 
  \label{fig:intro_ill}
\end{figure}

Despite their success, multi-view texturing still faces fundamental challenges as shown in Fig. \ref{fig:intro_ill}, including: (1) the lack of a robust inpainting scheme for occlusion regions; (2) the difficulty in achieving precise alignment of texture features with fine-grained geometric details; and (3) the challenge of maintaining multi-view consistency and coherence across content, color, and illumination. 
These errors can accumulate and manifest during the projection and baking process, introducing undesired artifacts to the textured results.
However, these challenges are inherently difficult to address due to several fundamental reasons. 
First, occlusion regions are an inevitable aspect of multi-view texturing; no matter the approach, they cannot be entirely avoided. 
Second, latent space diffusion inherently introduces errors, which makes pixel-level edge alignment hard to achieve, and 2D normal control is often insufficiently precise to handle fine-grained details. 
Third, maintaining consistency across multiple views is a costly process, and even state-of-the-art video models~\cite{kong2024hunyuanvideo,wan2025wan} struggle to achieve satisfactory results in this regard. 
Overall, these problems are broadly existing in 2D lifting methods~\cite{feng2025romantexdecoupling3dawarerotary,chen2023fantasia3d}, which are largely inevitable and stem from the cascading errors in modality changes. 

As a result, it remains a compelling yet significantly underexplored research direction: \emph{Can we treat 3D textures natively as first-class citizens to address the issues caused by modality change? What kind of paradigm would make this process more scalable?}
In a sense, treating 3D textures as first-class citizens naturally alleviates many of these challenges. 
Since textures are generated directly on the geometry surface, post-processing such as inpainting is no longer necessary. 
By directly injecting the entire geometry, the model avoids the inherent information loss caused by projecting 3D shapes into 2D views (\eg, depth or normal maps discard occluded regions and structural details). This allows the model to fully reason over the spatial context and achieve more accurate geometry–texture alignment.
Moreover, the coherent and unified representation of the entire 3D texture also simplifies the maintenance of global consistency. 
However, current solutions often build upon proxy representations (\eg, UV maps~\cite{liu2025texgarment, yu2023texture}, point-based features~\cite{xiong2025texgaussian, xiang2024structured}), resulting in data inefficiency as well as cascade errors. Thus, their ability to realize the full promise of 3D-native texture generation remains limited.

In this paper, we introduce \shortname, a novel {latent color diffusion model} that natively generates textures in 3D space. 
In contrast to previous approaches that rely on intermediate representations such as Gaussian Splatting~\cite{kerbl20233d,xiong2025texgaussian} or UV Maps~\cite{liu2025texgarment, yu2023texture}, \emph{\shortname directly predicts RGB color for given 3D coordinates via a latent diffusion approach}, a paradigm that has shown remarkable effectiveness in image~\cite{flux2024}, video~\cite{wan2025wan}, and 3D shape generation~\cite{lai2025hunyuan3d}, yet unexplored for texture generation. Specifically,  \shortname models texture as a dense color point cloud, effectively forming a continuous color field in 3D space. To mitigate the computational challenges of performing diffusion directly on a dense point cloud, we propose a color point cloud Variational Autoencoder (VAE) with a similar architecture to 3DShape2VecSet~\cite{zhang20233dshape2vecset}. Unlike 3DShape2VecSet, which focuses on shape autoencoding, our model operates on color point clouds. 
We retain the use of cross-attention for compressing the input point cloud into a set of latent vectors queried by points, while our set is \emph{ordered}, as the point queries are known and sampled from the input geometry at test time, which makes pointwise geometry condition possible.
We also introduce a new color regression loss that supervises both on- and near-surface regions.
Together, these adaptations yield an efficient autoencoder that achieves over 80× compression, enabling efficient scaling for subsequent diffusion transformer (DiT) generation.

Beyond occlusion-free and coherent representation for texture, another leading advantage of native texture models is \emph{native geometry control}, which greatly improves alignment. 
In contrast, previous multi-view texturing could only utilize fragmented geometric control, such as per-view normals and positions. This necessitated the design of complex consistency modules to maintain cross-view coherence. Meanwhile, many 3D structural details are inherently ambiguous when observed from a single 2D projection, making precise texture-geometry alignment difficult to achieve. 
In this work, we address these challenges through a novel native geometry control by co-designing the VAE–DiT architecture.
Our key idea is to integrate precise surface hints into DiT via pairwise conditional geometry tokens, implemented through positional embeddings and channel-wise concatenation.
To complement this, we design a dual-branch VAE that extends the color VAE with an additional geometry branch, encoding shape cues to guide color latent compression.
In this way, geometry tokens are deeply intertwined with color tokens, enabling stronger geometric guidance during color generation at test time.

Building upon the aforementioned designs, we further introduce a multi-control color DiT that flexibly accommodates different control signals.
Our design enables a wide range of applications beyond image-conditioned texture generation (with geometry and image controls) to texture-conditioned material generation and texture refinement (using an initial texture, named as color control).
Notably, \shortname exhibits remarkable generalization capability, enabling image-conditioned part segmentation and texturing even in a training-free manner.
To evaluate the effectiveness of our framework, we train \shortname-2B, which is mainly for texture generation but also adapted to previously mentioned applications for primary verifications. 
We report the comparison of \shortname-2B against previous methods, showing that it delivers exceptionally high-quality results and significantly outperforms prior approaches in texture oclusion, coherence and alignment.

Our main contributions are summarized as follows:
\begin{itemize}
\item We design a highly extensible framework for color field generation that, while demonstrated primarily on texture generation, can be readily extended to other tasks such as material modeling and part-level semantic segmentation.
\item We design a geometry-aware color VAE for color point clouds, in which a geometric branch encodes local shape cues to achieve geometric awareness for color generation.
\item We propose a novel multi-control color DiT that flexibly integrates geometric, image conditions, and color conditions, enabling seamless texture generation and beyond.
\item We achieve state-of-the-art performance against prior methods, particularly in texture coherence and alignment. 
\end{itemize}

\section{Related Works}

\subsection{3D Texturing via 2D Priors}
A prevalent approach in 3D texture synthesis involves adapting pre-trained 2D models such as the text-to-image diffusion models~\cite{rombach2022high} for the 3D domain. 
An influential line of research employs Score Distillation Sampling (SDS) \cite{poole2022dreamfusion} for iterative texture optimization \cite{chen2023fantasia3d, liang2024luciddreamer, lin2023magic3d, zhang2024dreammat}. While these methods can generate detailed textures, they are often computationally expensive and tend to suffer from 3D inconsistencies, commonly referred to as the “Janus problem” (multi-faced objects). Similarly, iterative mesh painting techniques~\cite{richardson2023texture, chen2023text2tex}, which generate textures by inpainting multiple camera views, also struggle with maintaining seamless and consistent results.

To address these consistency challenges, recent work has focused on Multi-view Diffusion (MVD) models~\cite{shi2023zero123++, liu2023zero, wang2023imagedream, li2024era3d, long2024wonder3d}. These models are explicitly trained to generate camera-consistent, object-centric images from text or image prompts. In texture generation, this paradigm is extended by conditioning the diffusion process on 3D geometry~\cite{feng2025romantexdecoupling3dawarerotary, zhang2024clay, he2025materialmvpilluminationinvariantmaterialgeneration, zeng2024paint3d}, often through rendered inputs like depth or normal maps. This geometric conditioning ensures that the generated views align with the underlying surface, thereby minimizing artifacts in the baked texture.

\begin{figure*}[t] 
  \centering
  \includegraphics[width=\linewidth]{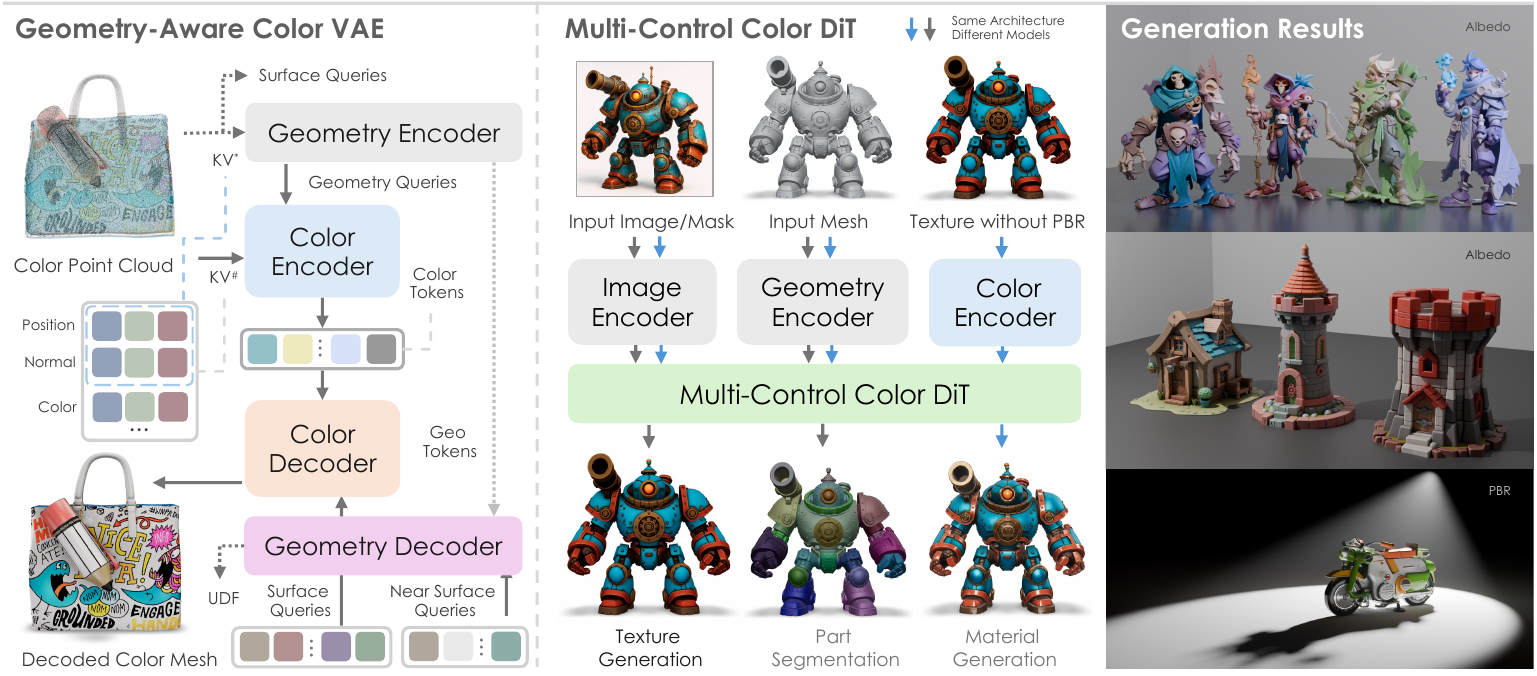}
  \caption{\textbf{Overall architecture of \shortname}: it mainly consists of a geometry-aware color VAE for reconstruction and a multi-control color DiT for generation, adaptable for diverse applications. Left-most assets are all generated by \shortname.} 
  \label{fig:arch}
\end{figure*}

\subsection{Native 3D Texture Generation}

Despite the success of MVD texturing in various commercial products~\cite{tripo,rodin,meshy,hitem3d,lai2025hunyuan3d}, these projection-based methods still struggle at keeping multi-view consistency and the precise alignment of texture with fine-grained geometric details. 
These problems could hardly be avoided as they are deeply intertwined with the inherent limitations of the baking process and MVD itself — unless one can instead generate textures directly and natively within 3D space.

In fact, early approaches to 3D asset creation, such as SDS~\cite{poole2022dreamfusion,chen2023fantasia3d,liu2024unidream} and large reconstruction models \cite{hong2023lrm,tang2024lgm,yang2024hunyuan3d} are native texture generators, though they typically generate geometry and texture simultaneously, and the quality is limited. Generally, previous attempts at native texture generation have focused on utilizing Generative Adversarial Networks (GAN)~\cite{siddiqui2022texturify,oechsle2019texture}, feed-forward~\cite{xiong2025texgaussian}, or diffusion models~\cite{liu2024texoct,yu2023texture,liu2025texgarment,yu2024texgen}, to predict face color~\cite{siddiqui2022texturify}, point color~\cite{liu2024texoct,yu2023texture}, UV color~\cite{liu2025texgarment,yu2024texgen, zeng2024paint3d,yu2023texture}, texture field~\cite{oechsle2019texture}, or Gaussian splatting~\cite{xiong2025texgaussian}.
Some methods, such as TexGaussian~\cite{xiong2025texgaussian} and TexOct~\cite{liu2024texoct}, adopt octree-representation for compression, while TexGarment~\cite{liu2025texgarment} adopts 2D VAE by viewing UV maps as images. More recently, Trellis~\cite{xiang2024structured} has also demonstrated how to generate textured assets in one phase through their SLAT representation. UniTEX~\cite{liang2025unitex} presents a refinement module but is limited at low resolution due to the complexity of triplane representation. 

In contrast to previous work, we focus on texture generation directly in 3D space, given an input geometry from artists or 3D generators~\cite{zhao2025hunyuan3d,lai2025flashvdm,lai2025hunyuan3d,hunyuan3d2025hunyuan3d,lei2025hunyuan3dstudioendtoendai,hunyuan3d2025hunyuan3domniunifiedframeworkcontrollable,yan2025x}. We demonstrate that 3D textures can be formulated within the scalable latent diffusion model paradigm—an approach not previously explored—yielding stunning results.

\section{Method}

It remains an open problem which generative paradigm offers the most scalable solution for texture synthesis.
Prior approaches that operate in view space via multiview texturing~\cite{zhao2025hunyuan3d,huang2024mvadapter, feng2025romantexdecoupling3dawarerotary} or directly generate color in UV space~\cite{yu2024texgen} often suffer from fundamental limitations, including misalignment with geometry, inability to handle occluded regions, and excessive reliance on UV unwrapping. 

The proposed \shortname is a generative latent color diffusion model that synthesizes textures directly in 3D space. As shown in Fig.~\ref{fig:arch}, it follows the standard latent diffusion architecture consisting of a geometry-aware color VAE~\cite{kingma2013auto} and a multi-control color
DiT~\cite{Peebles2022DiT}. 
In the following, we detail our representation of 3D texture, the designs of VAE, DiT, and the conditioning mechanisms, as well as the broader applications of our model.

\subsection{Color Representation and VAE}

Instead of modeling textures in projective 2D image space or UV space as in prior works~\cite{zhao2025hunyuan3d,he2025materialmvpilluminationinvariantmaterialgeneration,yu2024texgen}, we propose to represent textures natively in 3D as a color field. Concretely, the goal is to predict RGB values conditioned on geometric positions. To realize this, we leverage dense color point clouds sampled from textured meshes as our representation. Compared to view-space methods, our approach operates directly in 3D, naturally handling occluded regions without requiring inpainting. Compared to UV-space methods, it avoids reliance on UV quality and instead provides a more structured and coherent representation, better suited for generative modeling.

\textbf{Geometry-Aware Autoencoding.} Insipred by the success of native geometry generation~\cite{zhang2024clay,zhao2025hunyuan3d, hunyuan3d2025hunyuan3d}, we adopt a VAE architecture similar to 3DShape2VecSet~\cite{zhang20233dshape2vecset} to encode color point cloud. While alternative designs are possible, we leave their exploration to future work. A visual overview of the architecture is provided in Fig.~\ref{fig:arch}. 
The VAE takes as input a point cloud $P_c \in \mathbb{R}^{N\times9}$ sampled from a textured mesh, containing RGB color, position, and normal, and reconstructs a continuous color field $f(\mathbf{x}) = \mathbf{c}$, mapping each 3D coordinate $\mathbf{x} \in \mathbb{R}^{3}$ to its color $\mathbf{c} \in \mathbb{R}^{3}$.

One of the unique problems for texture generation is how to incorporate fine-grained geometric conditioning during generation. A straightforward solution would be to encode geometry using a VecSet-based ShapeVAE~\cite{zhang20233dshape2vecset,zhao2025hunyuan3d} on the same color point cloud. 
Instead, we propose a tighter integration, as illustrated in Fig.~\ref{fig:arch}. In parallel with the texture VAE, we introduce a geometry VAE branch that encodes geometric features from the coordinates and normals of the color point cloud. The resulting geometry latent set is then employed as queries to guide the texture encoder.
Concretely, the input point cloud is uniformly sampled and consists of positions, normals, and colors. The geometry encoder consumes positions and normals, while the color encoder leverages all three modalities. Geometry queries are constructed as point queries~\cite{zhang20233dshape2vecset} randomly sampled from the color point cloud. For both geometry and texture encoding, we adopt the same network backbone as Hunyuan3D-VAE~\cite{zhao2025hunyuan3d}, which incorporates multiple layers of cross-attention and self-attention.

During the inference, our model supports two common output modalities: it can synthesize a UV texture map by mapping UV coordinates to 3D and querying colors via 3D coordinates, or it can directly predict per-face / per-vertex colors by querying at vertex or face-center coordinates. Similar to geometry VAE~\cite{zhao2025hunyuan3d}, the texture VAE operates at arbitrary resolution, enabling flexible decoding for different downstream targets.
During training, the geometry and texture VAEs are jointly optimized with a KL divergence term, a color regression loss that supervises both on-surface and near-surface queries, and a truncated UDF loss:
\begin{equation}
o(\mathbf{x}) = 
\begin{cases} 
1, & \text{if } udf(\mathbf{x}) > s \\
\dfrac{udf(\mathbf{x})}{s}, & \text{if } udf(\mathbf{x}) \leq s ,
\end{cases}
\end{equation}
where the truncated UDF is adopted because correlating the color point cloud with a watertight mesh (required for standard SDF loss) is non-trivial.
For the color regression loss, we supervise both on-surface points and near-surface points. The latter is implemented by randomly offsetting query points along their normal directions within a threshold $\gamma$.
The overall training objective is thus:
\begin{equation}
\mathcal{L} = \lambda_{\text{KL}}\mathcal{L}_{\text{KL}} + \lambda_{\text{color}}\mathcal{L}_{\text{color}} + \lambda_{\text{udf}}\mathcal{L}_{\text{UDF}}.
\end{equation}
where $\lambda_{\text{KL}}, \lambda_{\text{color}}, \lambda_{\text{udf}}$ are loss weights.

\textbf{Reconstruct any Field.} A key advantage of our autoencoding framework is its universality: any modality that admits an RGB-like representation can be seamlessly incorporated into the same latent space. For example, physically based materials can be encoded using a unified color VAE by mapping metallic and roughness into a modified albedo, where the blue channel is fixed to zero. Likewise, semantic part segmentation can be treated in the same way by mapping discrete part labels into color values. This unified treatment allows diverse signals—ranging from appearance to semantics—to be represented and processed within a single, coherent framework.

\begin{figure}[t] 
  \centering
  \includegraphics[width=\linewidth]{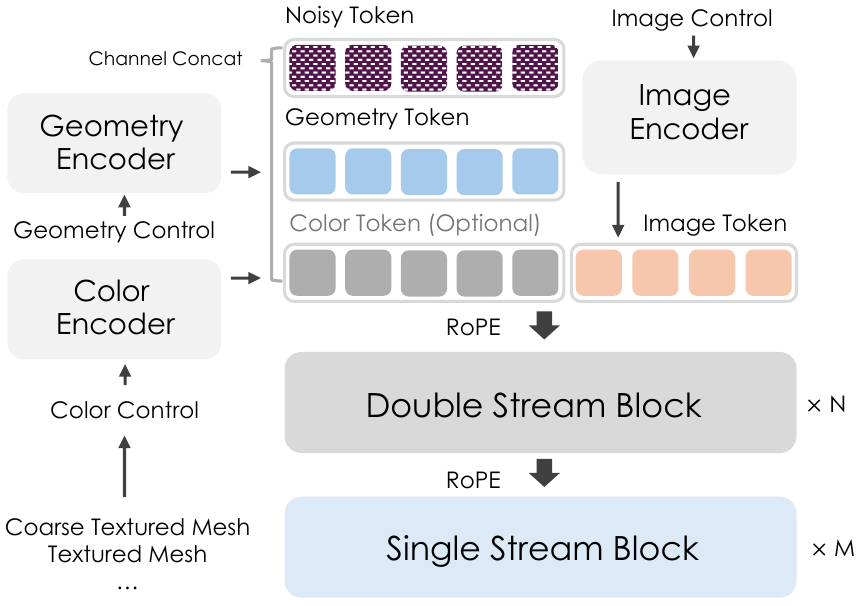}
  \caption{Illustration of multi-control mechanisms of the proposed color DiT. Color control is useful for texture-conditioned tasks.} 
  \label{fig:cond_dit}
\end{figure}

\subsection{Multi-Control Color DiT}
\label{sec:dit}

We adopt an architecture similar to the rectified flow diffusion transformer~\cite{flux2024} for generating the texture latent set. To accommodate richer control signals, we introduce several adaptations that allow the model to incorporate not only the input image but also the input geometry and an initial texture, as shown in Fig.~\ref{fig:cond_dit}.

\textbf{Image Control.}
Following Hunyuan3D-2~\cite{zhao2025hunyuan3d}, we use Dinov2-Giant~\cite{oquab2023dinov2} for image conditioning, utilizing the embedding from the last hidden layer without the class token. Unlike Hunyuan3D-2~\cite{zhao2025hunyuan3d}, which uses a resolution of 518, we scale the input to 1022, as we found higher-resolution conditioning helps for capturing fine-grained details. To minimize image token length, we retain the original aspect ratio by cropping the object from the 1022-resolution input image using a binary mask. No additional positional embedding is used for image tokens, as we find the position information encoded in Dino to be sufficient.

\begin{figure*}[t] 
  \centering
  \includegraphics[width=\linewidth]{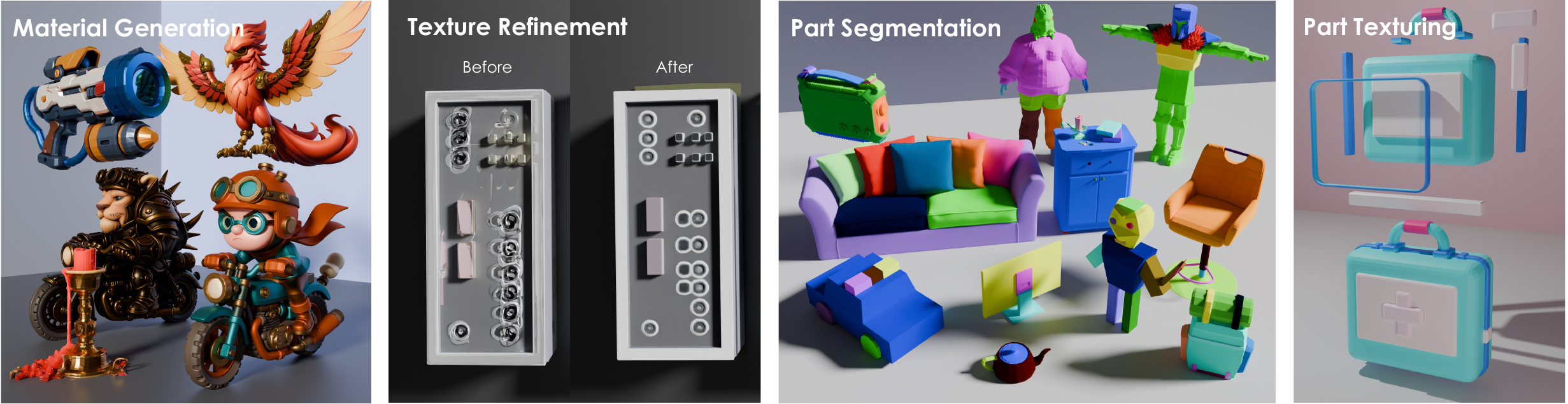}
  \caption{Visual results showcasing representative applications of \shortname. Additional results are provided in the Appendix.} 
  \label{fig:application}
\end{figure*}
\begin{figure}[t] 
  \centering
  \includegraphics[width=\linewidth]{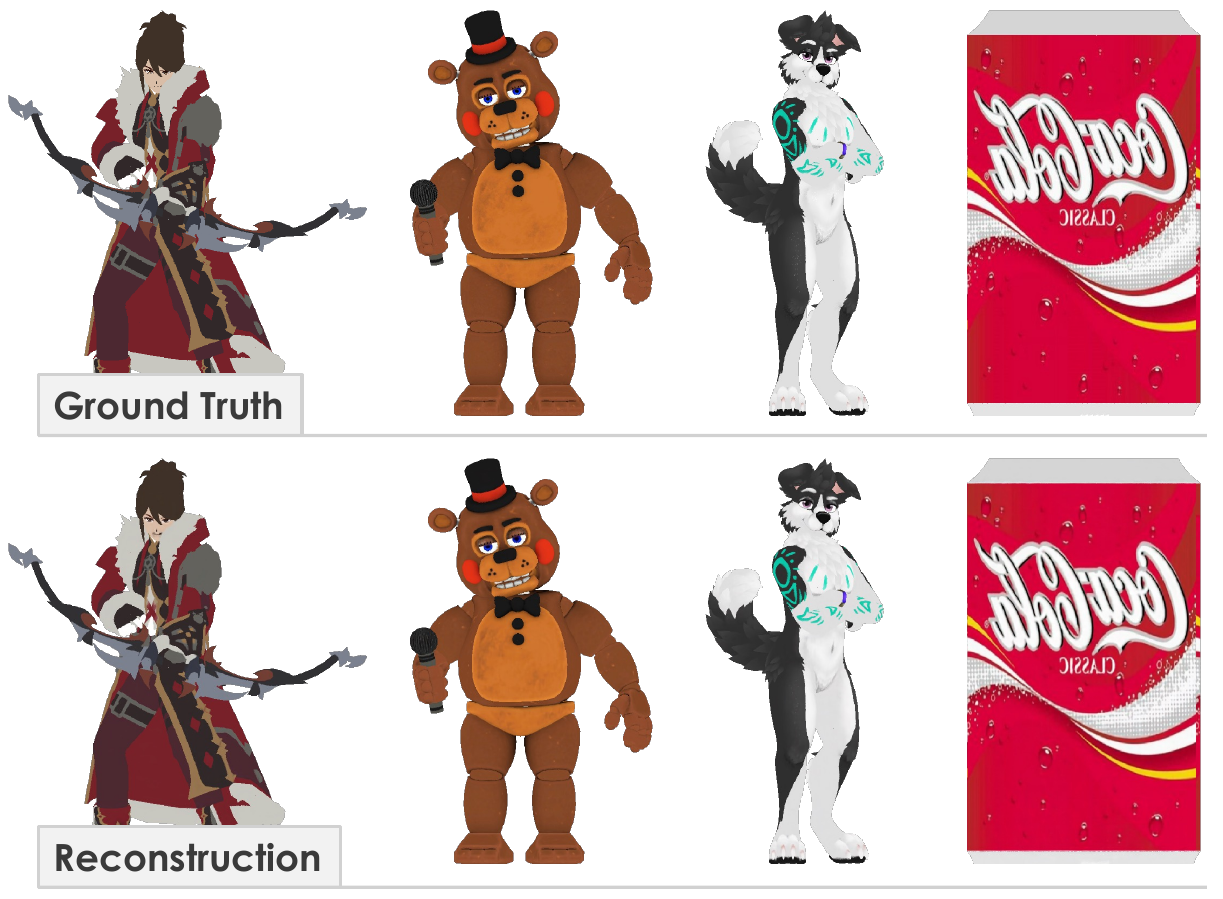}
  \caption{Visual comparison of texture reconstruction results.} 
  \label{fig:reconstruct}
\end{figure}

\textbf{Geometry Control.} Geometry conditioning plays a crucial role in aligning the generated texture with the input mesh. In this work, we propose \emph{native geometry control}, which includes two conditions: (1) we incorporate RoPE~\cite{su2024roformer} based on the positions of sampled point queries, which provides coarse structural guidance; and (2) we leverage the geometry latent set obtained from the VAE (described in the previous section) as an additional embedding to deliver fine-grained guidance. Furthermore, since the geometry latent set is isomorphic to the texture latent set, we concatenate it with the noisy texture latent set along the channel dimension, enabling pointwise geometric guidance during texture generation.

\textbf{Color Control.}
Our model also supports incorporating an initial texture as extra control (termed color control), which is useful for various downstream tasks, such as texture-conditioned material generation as well as texture inpainting and refinement. To achieve it, we sample a color point cloud from the given texture and encode it using our VAE to obtain a conditional color latent set. This latent set is then concatenated with the noisy texture latent and geometry latent along the channel dimension, providing stronger guidance while keeping the sequence length unchanged.

\textbf{Training and Inference Strategies.} 
During training, we encode the sampled color point cloud using our VAE to obtain aligned geometry and texture latent sets. Each latent token is associated with a position, which is used for RoPE~\cite{su2024roformer} applied to the noisy texture latent. The model is trained with a flow matching~\cite{lipman2022flow} loss. For albedo generation, following MaterialMVP~\cite{he2025materialmvpilluminationinvariantmaterialgeneration}, we also include an illumination-invariant loss, which results a hybrid loss:
\begin{equation}
    \mathcal{L} =  ||\epsilon_{pred}-\epsilon_{gt}||_2^2 + \gamma  ||\epsilon_{pred}-\epsilon_{pred2}||_2^2
\end{equation}
where $\epsilon_{pred2}$ and $\epsilon_{pred}$ are the predictions for input images with different illuminations.  
During inference, we first convert UV coordinates or vertex positions into a point cloud and sample normals from the input geometry. The geometry point cloud is then encoded by the geometry branch of our VAE, producing a geometry latent set along with corresponding latent positions. These geometry conditions, together with the input image, are fed into the generator for a diffusion sampling step to produce the final texture.

\subsection{Applications}
\shortname provides a unified framework that is broadly applicable across diverse 3D tasks, as shown in Fig.~\ref{fig:application}. Below, we highlight several representative use cases:

\textbf{Material Generation.}  
Physical-based rendering becomes extremely straightforward by treating roughness and metallic as two channels in an RGB color point cloud. By leveraging the same color VAE framework, we can train a generator conditioned on an additional albedo latent set extracted from the proposed color VAE, facilitating efficient material synthesis with high fidelity.

\textbf{Texture Refinement.} Our model with color control can be viewed as a neural refiner that automatically inpaints occluded regions and corrects texture. Moreover, thanks to the strong conditioning, it can perform the process in just five steps, making it extremely fast and efficient for a wide range of downstream tasks that require intelligent refinement.

\textbf{Part Segmentation.}  
Without any additional training, our model can be easily adapted for part segmentation tasks. This can be achieved by feeding 2D segmentation of the input image into the model, allowing it to generate a texture map that aligns with 3D part segmentation results.

\textbf{Part Texturing.} Within our native texture framework, part generation is as straightforward as generating textures for the entire object, as we can predict color directly in 3D space for different part surfaces. Moreover, our method naturally handles occluded regions, producing consistent and clean textures for different parts.

\begin{figure*}[t] 
  \centering
  \includegraphics[width=\linewidth]{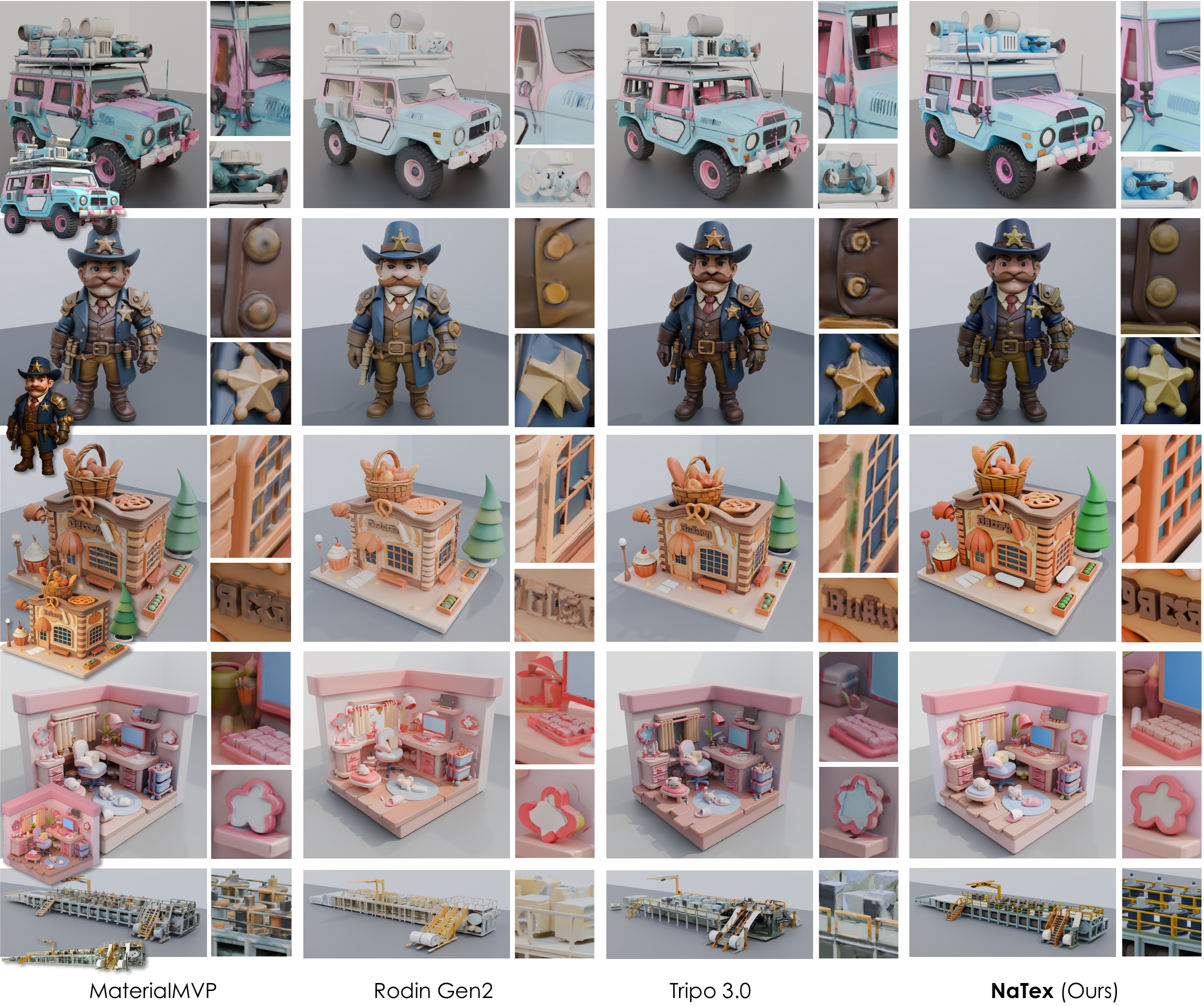}
  \caption{Visual comparison of different methods for generating textured 3D assets from images: commercial models use their own geometries, while other methods share the same geometry from Hunyuan3D 2.5~\cite{lai2025hunyuan3d}. All methods are rendered with albedo only.} 
  \label{fig:gen_compare}
\end{figure*}

\begin{table}[t]
    \centering
    \setlength{\tabcolsep}{6.5pt}
    \small
    \begin{tabular}{cccccc}
    \toprule 
    \textbf{Latent Size}     & \textbf{PSNR$\uparrow$} & \textbf{PSNR$\uparrow^{*}$} & \textbf{SSIM$\uparrow^{*}$} & \textbf{LPIPS$\downarrow^{*}$}  \\ \midrule
   $6144\times 64$   & 28.74 & 31.70 & 0.980  & 0.0492 \\
   $12288\times 64$  & 29.95 & 33.19 & 0.984  & 0.0445 \\
   $24576\times 64$  & 30.86 & 34.30 & 0.987  & 0.0411       \\
    \bottomrule
    \end{tabular}
    \caption{Quantitative results for texture reconstruction. $^*$ denotes metrics calculated on the six orthogonal rendered views.}
    \label{tab:texture_vae}
\end{table}

\section{Experiments}

\subsection{Comparison}

\textbf{Reconstruction.} 
To the best of our knowledge, we are the first method to utilize a native latent diffusion model for texture generation. We evaluate the reconstruction performance of our method across different latent sizes. We adopt several metrics for evaluation: PSNR is computed directly on the color point cloud, while  PSNR$^*$ and SSIM$^*$~\cite{wang2004image} and LPIPS$^*$~\cite{zhang2018unreasonable} are calculated by rendering the reconstructed textured meshes into 2D images.
The numerical comparison is shown in Tab.\ref{tab:texture_vae}. Although our model is trained with a maximum of 6144 tokens, the reconstruction quality improves as the latent size increases. The visual comparison of reconstruction is shown in Fig.\ref{fig:reconstruct}.

\textbf{Generation.}
We perform a quantitative comparison with other image-conditioned texture generation methods, including Paint3D~\cite{zeng2024paint3d}, TexGen~\cite{yu2024texgen}, Hunyuan3D-2~\cite{zhao2025hunyuan3d}, RomanTex~\cite{feng2025romantexdecoupling3dawarerotary}, and MaterialMVP~\cite{he2025materialmvpilluminationinvariantmaterialgeneration}. The comparison focuses solely on albedo results.
Following the evaluation protocol from MaterialMVP~\cite{he2025materialmvpilluminationinvariantmaterialgeneration}, we use the same test set and four metrics for assessment: CLIP-based FID (c-FID), Learned Perceptual Image Patch Similarity (LPIPS), CLIP Maximum-Mean Discrepancy (CMMD), and LIP-Image Similarity (CLIP-I).
The numerical comparison is presented in Tab. \ref{tab:texture_gen}, where our method consistently outperforms the others.
Additionally, we provide a visual comparison with closed-source commercial models, Rodin-Gen2 and Tripo 3.0, in Fig. \ref{fig:gen_compare}. It is evident that all competing methods struggle to align textures along geometry boundaries, while our method achieves near-perfect alignment. Notably, even in cases without occlusion regions (\eg, the character), other methods still produce artifacts, such as misaligned stars and buttons.

\begin{figure}[t] 
  \centering
  \includegraphics[width=\linewidth]{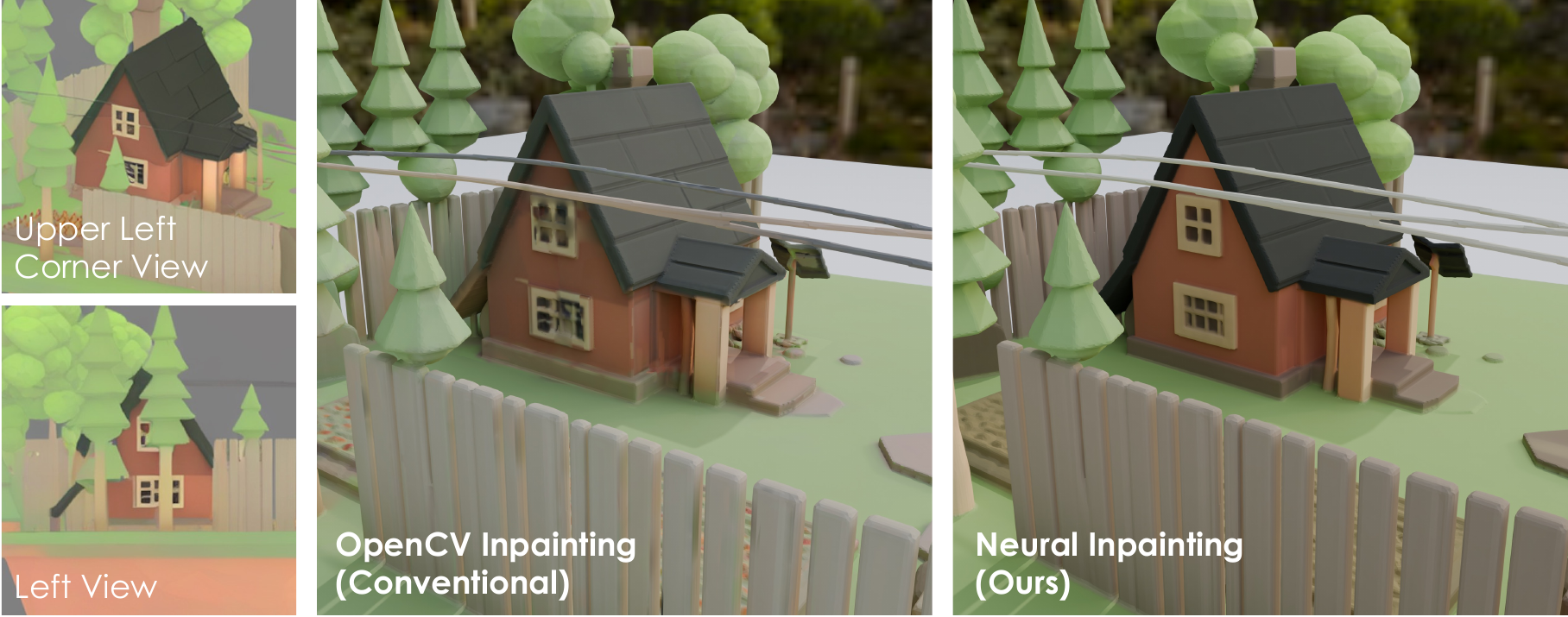}
  \caption{Visual comparison of conventional inpainting and our neural inpainting. Two views of multiview images are shown on the left. We need to inpaint the occlusion in the window.} 
  \label{fig:inpaint}
\end{figure}
\begin{figure}[t] 
  \centering
  \includegraphics[width=\linewidth]{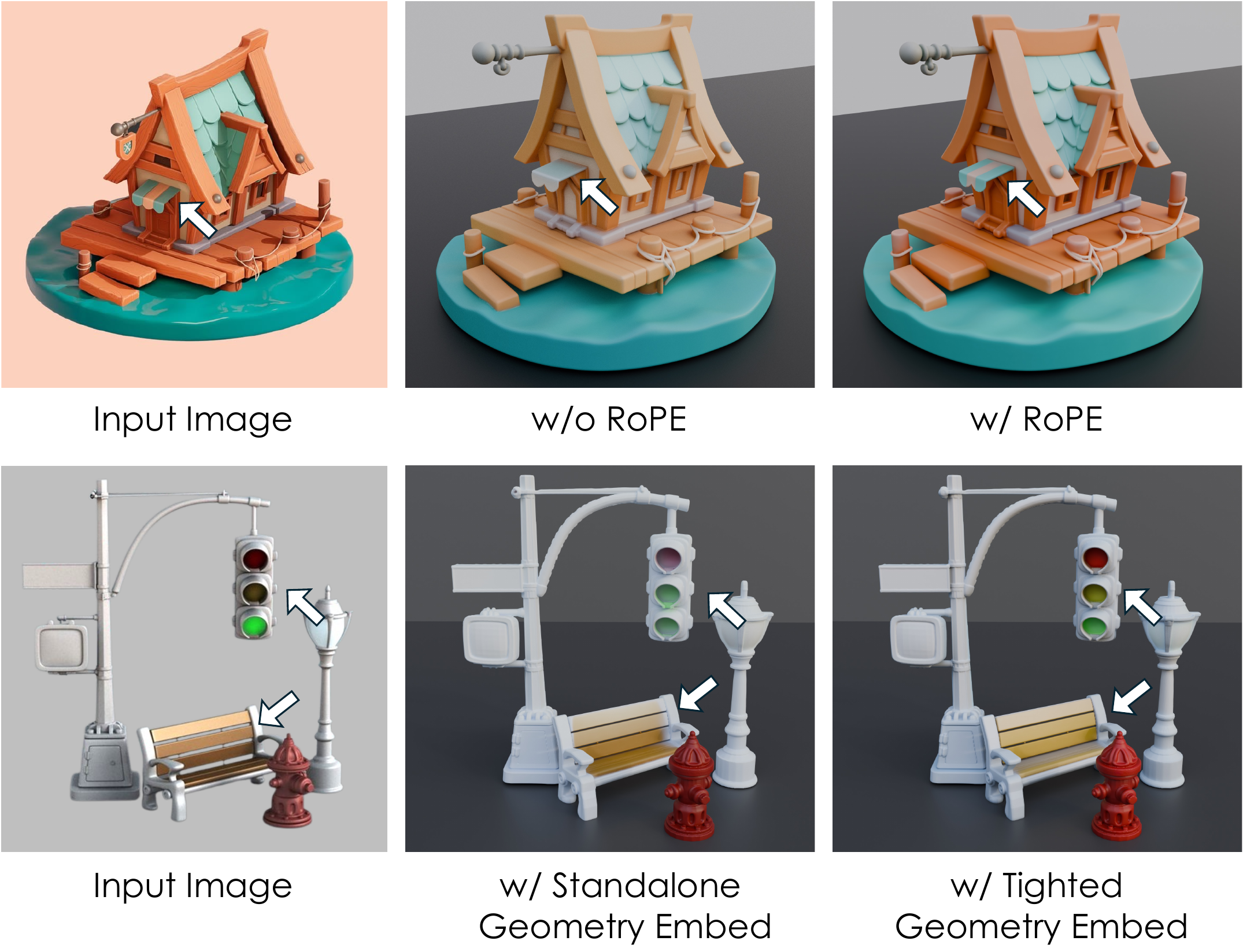}
    \vspace{-6mm}
  \caption{Visual comparison of different geometry conditioning methods: with RoPE and geometry embedding from the geometry-aware color VAE, texture and color alignment are optimized.} 
  \label{fig:ablation}
\end{figure}
\begin{figure}[t] 
  \centering
  \includegraphics[width=\linewidth]{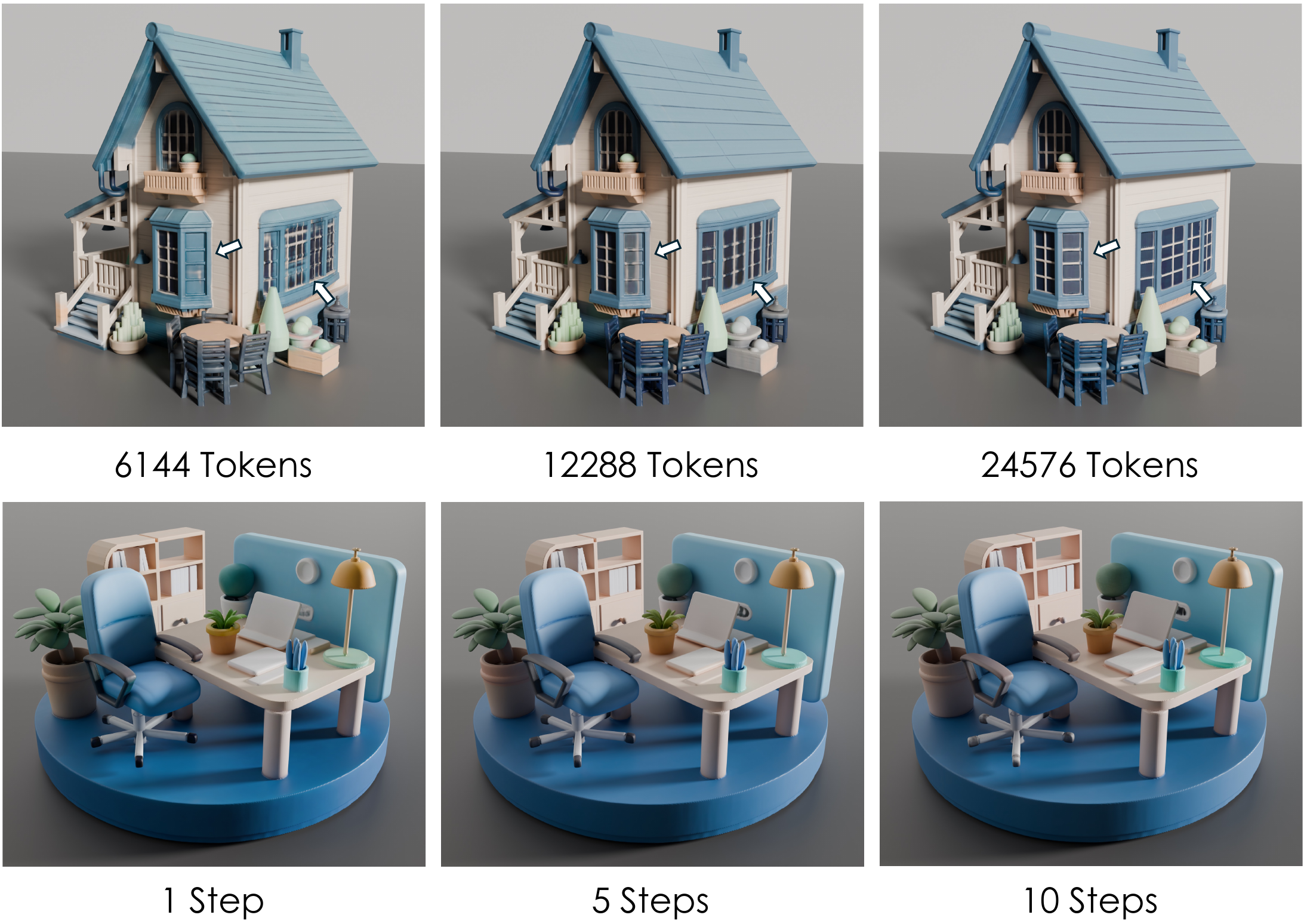}
  \vspace{-6mm}
  \caption{Illustration of different inference setups. Without additional training and distillation, more tokens improve quality and alignment, and we achieve one-step generation for free.} 
  \label{fig:ablation_infer}
\end{figure}

\textbf{Refinement/Inpainting.} Our model can also function as a refinement or inpainting module by utilizing the color control proposed in Sec. \ref{sec:dit}. We compare our method with the conventional inpainting technique, which uses OpenCV interpolation. As shown in Fig. \ref{fig:inpaint}, it is clear that our method generates cleaner and better-aligned textures in the occluded regions (see the zoomed-in window of the house).

\subsection{Evaluation}

\textbf{Ablation Study of Training Strategies.} 
We evaluate different setups for the proposed color DiT with varying geometry conditions. The first variant removes the RoPE from each color token, while the second variant replaces the tight shape embedding from the proposed geometry-aware color VAE with a shape embedding from a standalone shape VAE, such as Hunyuan3D-VAE~\cite{zhao2025hunyuan3d}. The comparison is shown in Fig.~\ref{fig:ablation}, where we observe that both conditions improve image-texture alignment, such as the stripes on the awning of the house and the colors of the traffic light. Additionally, the shape embedding influences texture-geometry alignment. Without the tight embedding, color sometimes diffuses, as seen in the chair back.

\begin{table}[t]
    \centering
    \setlength{\tabcolsep}{5pt}\small
    \begin{tabular}{ccccccc}
    \toprule 
    \textbf{Method}     &  \textbf{cFID$\downarrow$} & \textbf{CMMD$\downarrow$} & \textbf{CLIP$\uparrow$} & \textbf{LPIPS$\downarrow$}  \\ \midrule
    Paint3D \cite{zeng2024paint3d}     & 26.86      & 2.400 & 0.887      & 0.126   \\
    TexGen \cite{yu2024texgen}      & 28.23      & 2.447 & 0.882 & 0.133     \\ 
    Hunyuan3D-2 \cite{zhao2025hunyuan3d}     & 26.43     & 2.318 & 0.889 & 0.126       \\ 
    RomanTex \cite{feng2025romantexdecoupling3dawarerotary}     & 24.78      & 2.191 & 0.891 & 0.121       \\ 
    MaterialMVP \cite{he2025materialmvpilluminationinvariantmaterialgeneration}      & {24.78}      & {2.191} & \textbf{0.921} & {0.121}       \\ 
    \hline
    \textbf{\shortname} (Ours)       & \textbf{21.96}      & \textbf{2.055} & {0.908} & \textbf{0.102}       \\ \bottomrule
    \end{tabular}
    \caption{Quantitative comparison with state-of-the-art methods. }
    \label{tab:texture_gen}
\end{table}

\textbf{Effect of Different Inference Schemes.} 
Although our model is trained with a maximum of 6144 tokens, it supports various inference schemes at test time. Fig.~\ref{fig:ablation_infer} demonstrates the results under different token lengths and sampling steps. It can be observed that both generation quality and alignment improve gradually as the token length increases (see windows). Moreover, our model surprisingly achieves one-step generation capability, despite not being distilled, due to the strong conditioning.

\section{Conclusion}

In this paper, we introduce \shortname, a novel framework for generating textures directly in 3D space. We demonstrate that 3D texture generation can be formulated as simply as common latent diffusion, an extremely successful paradigm in image, video, and geometry generation, without the need for multi-stage pipelines with 2D lifting priors. Through careful design of both the VAE and DiT, our method effectively mitigates several key challenges, such as texturing occlusion regions, fine-grained texture-geometry alignment, and texture consistency—issues that have been inherently difficult to address in previous methods. 
Additionally, our model exhibits strong versatility across a wide range of downstream tasks, even without any task-specific training. 

\appendix

\section{Implementation Details}
\textbf{Training Details.} 
To validate the proposed method, we train a color VAE with 300M parameters and a color DiT with 1.9B parameters using a flow-matching objective. The VAE is trained with a maximum of 6144 tokens, with token scaling during inference. For DiT training, we set the batch size to 256 and use a constant learning rate scheduler with a linear warm-up for the first 500 steps. The learning rate starts at $1 \times 10^{-4}$ and decays to $1 \times 10^{-5}$ thereafter. The illumination-invariant loss is introduced once pretraining converges, with a weight of 5.
We adopt classifier-free guidance~\cite{ho2022classifier} by replacing conditioning embeddings with zero embeddings at a 10\% probability during training.
Unless otherwise stated, all results in this paper are obtained with 5 diffusion steps and a guidance scale of 2. The illumination-invariant loss is introduced once pretraining converges, with its weight set to 5.

\textbf{Data Preparation.}
We use Blender to sample uniform color point clouds from raw meshes. For the input images, we render 24 views uniformly around the object, with random elevation angles in the range of 45° to -30°. We also randomly select from various illumination environments, including point lights, area lights, and HDRI maps.

\begin{figure}[t] 
  \centering
  \includegraphics[width=\linewidth]{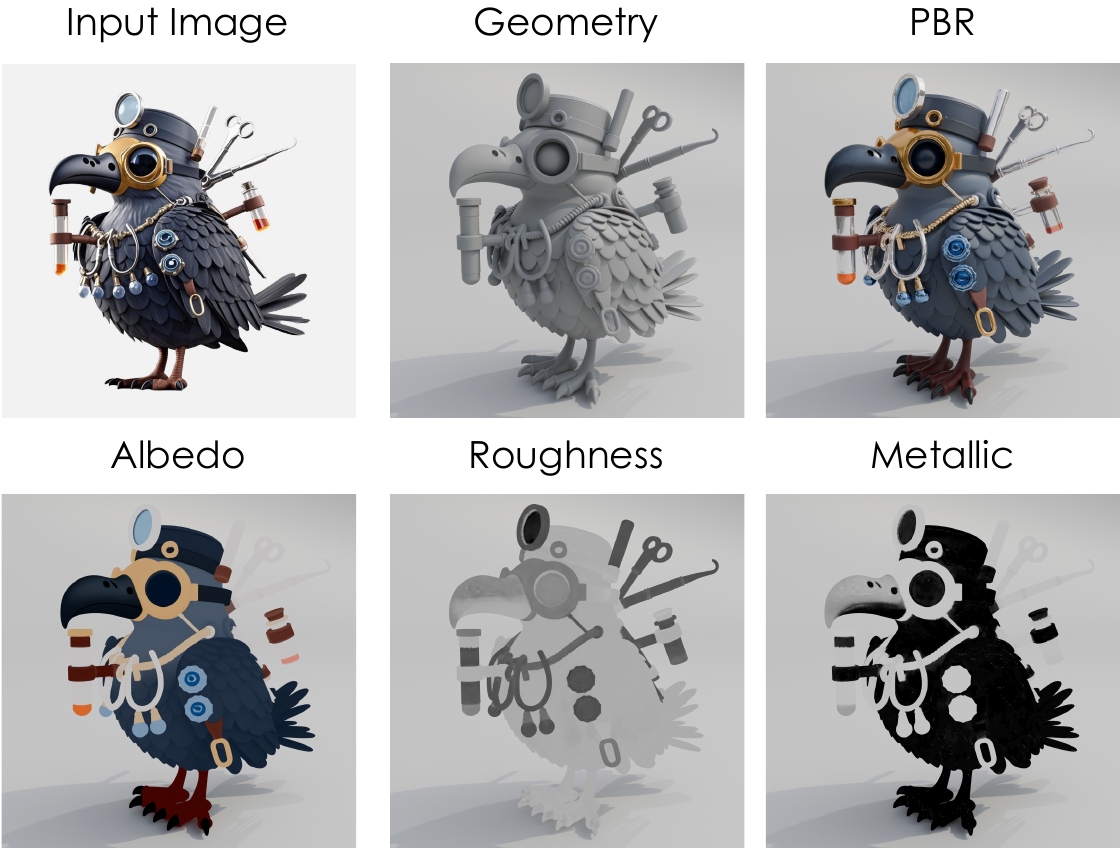}
  \caption{Illustration of our material generation results from a case study, with individual components visualized separately.} 
  \label{fig:material_case_study}
\end{figure}

\begin{figure}[t] 
  \centering
  \includegraphics[width=\linewidth]{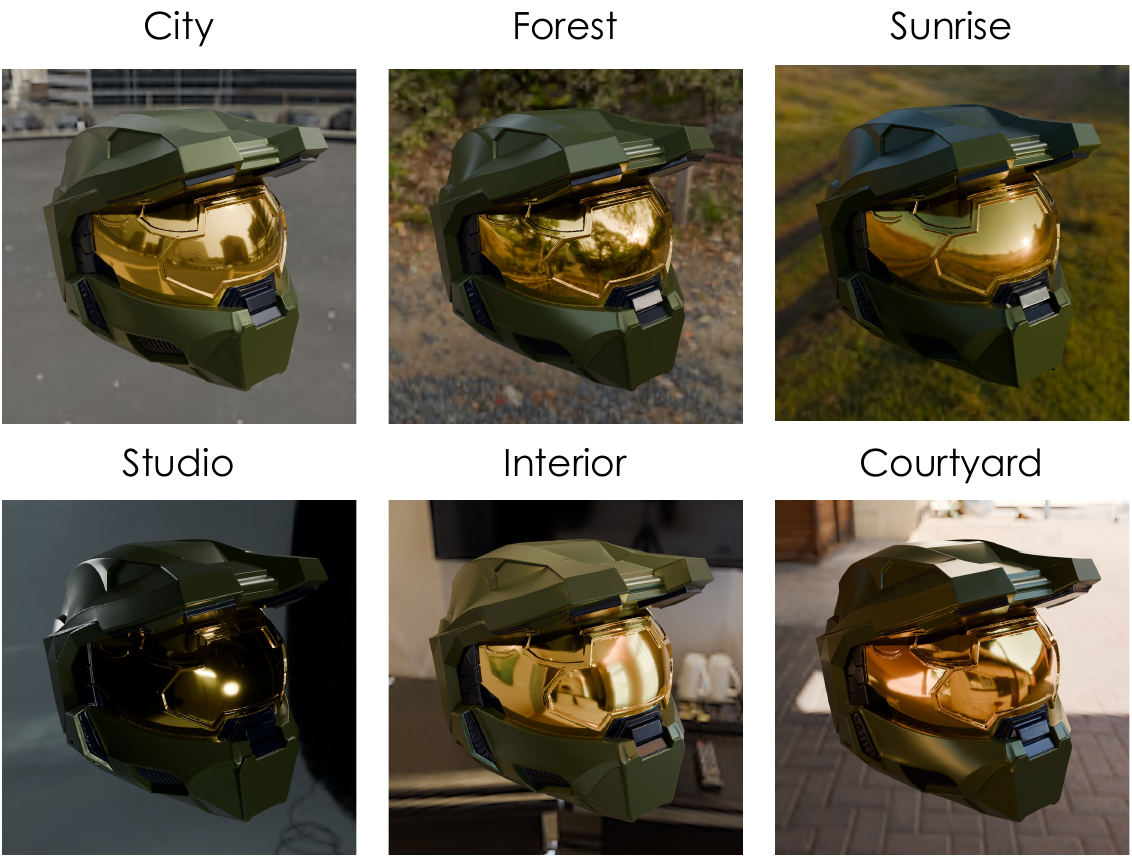}
  \caption{Illustration of our material generation results under different lightings, rendered using various environment maps.} 
  \label{fig:material_envs}
\end{figure}

\begin{figure}[t] 
  \centering
  \includegraphics[width=\linewidth]{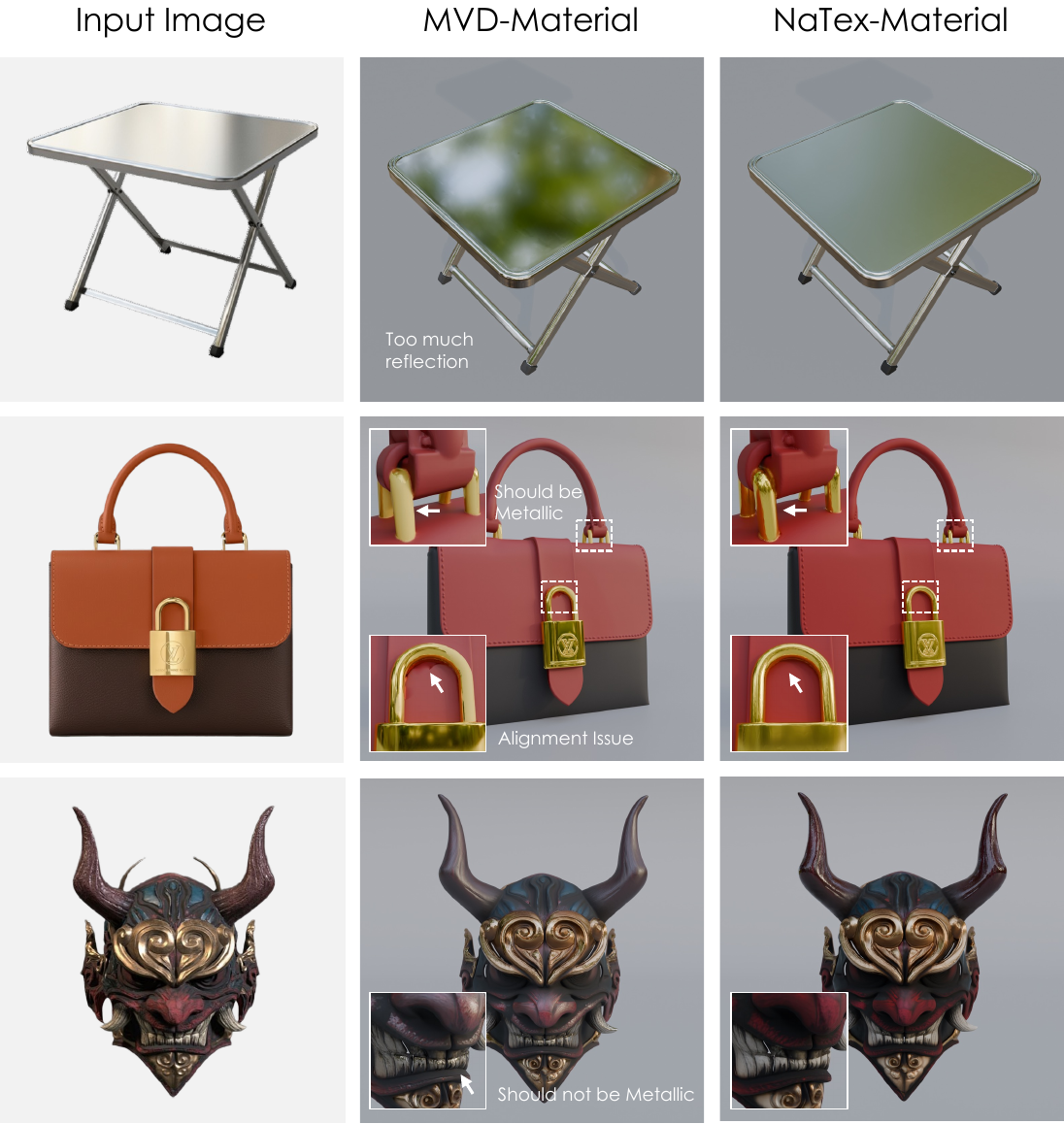}
  \caption{Visual comparison between our NaTex material generation pipeline and a conventional MVD-based material pipeline. Our method produces more accurate and better-aligned materials compared to prior approaches.} 
  \label{fig:material_cmp}
\end{figure}

\begin{figure*}[t] 
  \centering
  \includegraphics[width=\linewidth]{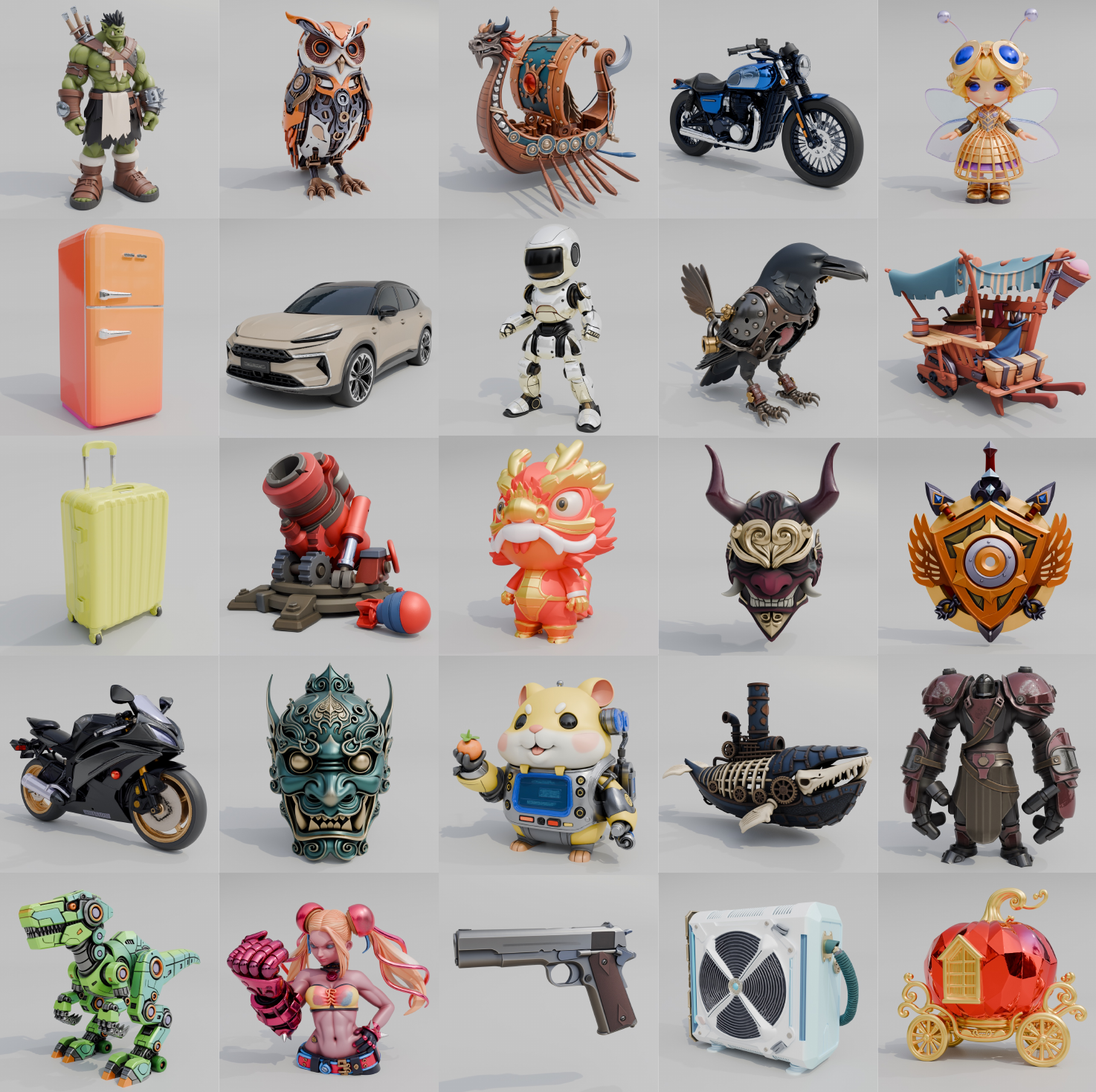}
  \caption{High-quality PBR-textured assets generated by NaTex. Geometry obtained from Hunyuan3D 2.5~\cite{lai2025hunyuan3d}.} 
  \label{fig:material_good}
\end{figure*}

\section{More Details on Applications}

\textbf{Material Generation.}  
Thanks to the flexible design of the proposed NaTex framework, we can easily adapt it for material generation with color control. Specifically, we formulate material generation as a two-channel texture generation task conditioned on the textured mesh with albedo. We reuse the same color VAE employed for texture generation, representing roughness and metallic as two channels in an RGB color point cloud. A new material DiT is then trained on this material color point cloud data, conditioned on the input image (image control), the textured mesh with albedo (color control), and the input geometry (geometry control). During inference, we adopt a two-stage approach: the first stage predicts the albedo, and the second stage predicts roughness and metallic based on the previously predicted albedo.

The generation results of NaTex-Material inherit the advantages of native texture generation, producing well-aligned and coherent roughness and metallic maps, as shown in Fig.\ref{fig:material_case_study}. We believe this represents a significant advantage for developing next-generation material generation frameworks, since previous MVD approaches often struggle with alignment and sometimes misinterpret material properties, as illustrated in Fig.\ref{fig:material_cmp}.

\begin{figure*}[t] 
  \centering
  \includegraphics[width=\linewidth]{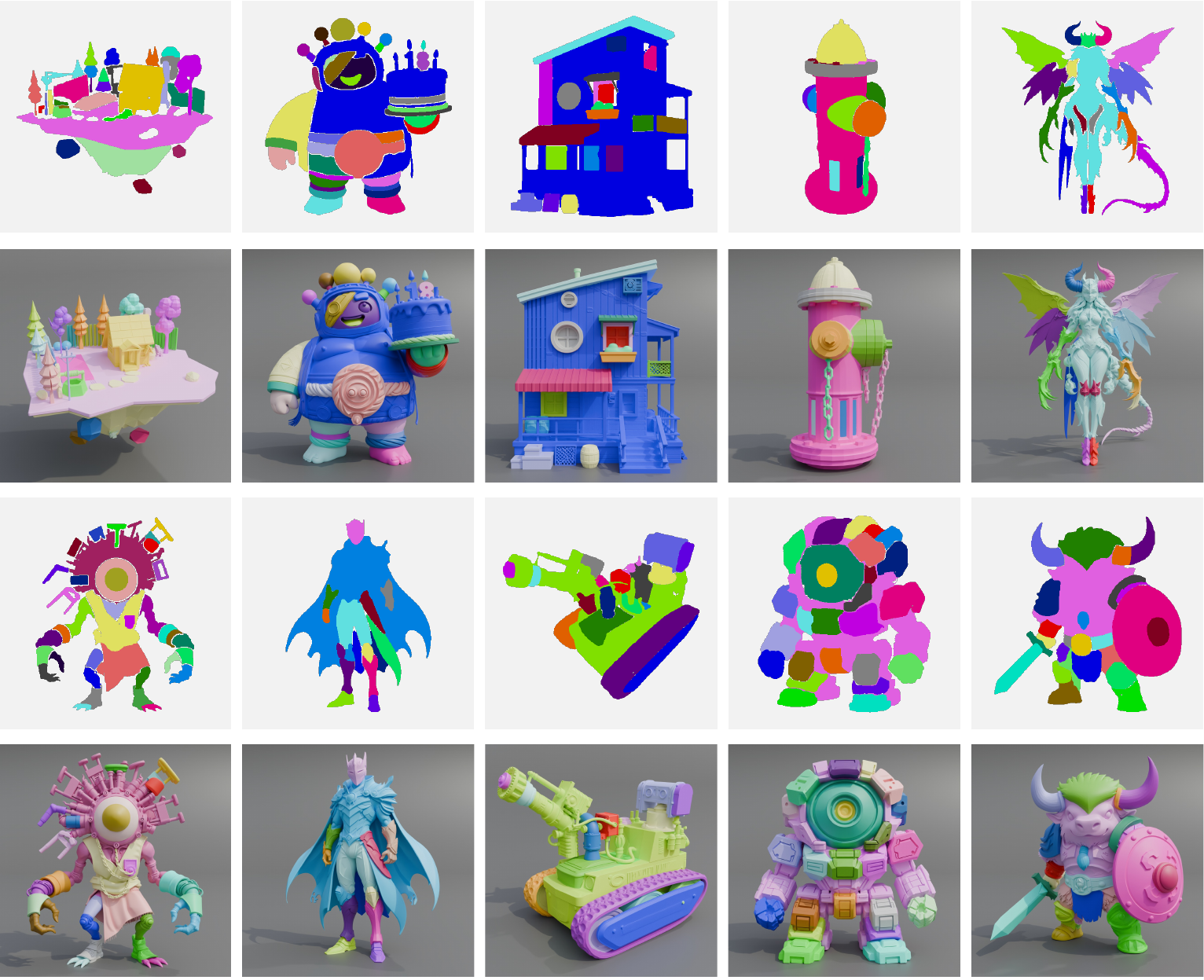}
  \caption{Visual results of part segmentation using a finetuned version of NaTex-2B. We provide a 2D mask as the input image for the given geometry, and NaTex textures the model accordingly.} 
  \label{fig:part_seg_good}
\end{figure*}

Fig.\ref{fig:material_envs} presents our material generation results under different lighting conditions, demonstrating the effectiveness of the generated materials. Fig.\ref{fig:material_good} showcases additional high-quality PBR-textured assets generated by NaTex, with albedo, roughness, and metallic maps all produced natively by our framework.

\textbf{Part Segmentation.}
We find that our model can be readily applied to part segmentation by conditioning on a 2D mask, as indicated in the main paper. Specifically, this can be achieved by first performing semantic segmentation on the input RGB image using SAM\cite{kirillov2023segment}. We then directly apply our texture model, NaTex-2B, without any additional training, feeding in the 2D mask to obtain the textured mesh. 

Nevertheless, this zero-shot strategy may produce fragmented or inconsistent results for complex structures. To address this, we finetune the base model on a dedicated dataset.
Surprisingly, the results of the finetuned model are highly accurate even on complex cases, as shown in Fig.\ref{fig:part_seg_good}, providing strong 3D segmentation with well-aligned boundaries. This further demonstrates the effectiveness and adaptation capability of our model.

\begin{figure}[t] 
  \centering
  \includegraphics[width=\linewidth]{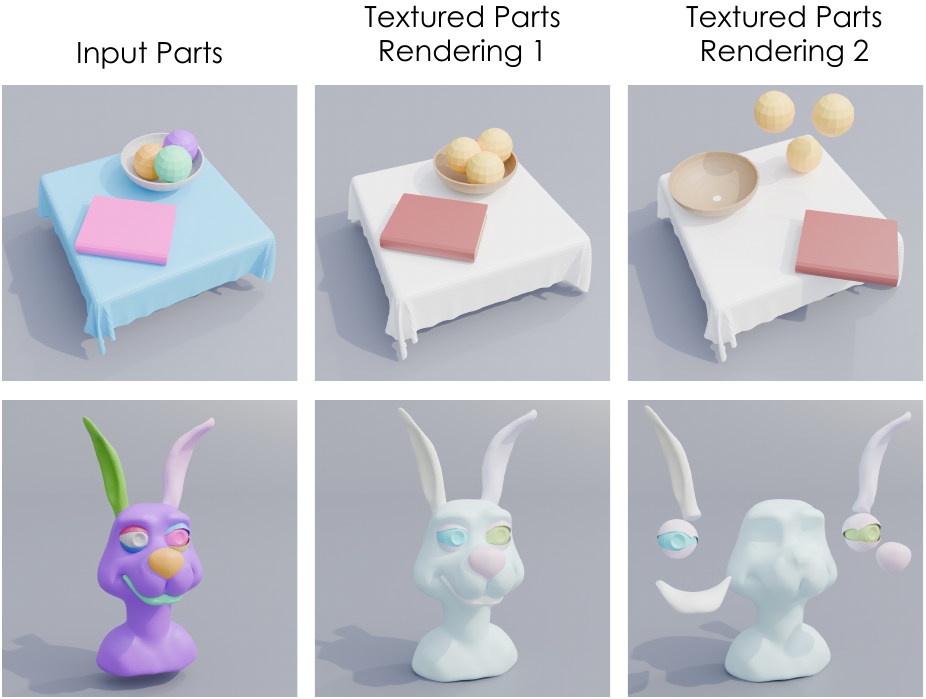}
  \caption{Illustration of part texturing using NaTex without any additional training. Our model generates textures for different parts without suffering from occlusion issues between them, as shown in the two renders with varying part arrangements.} 
  \label{fig:part_texture_case}
\end{figure}

\begin{figure}[t] 
  \centering
  \includegraphics[width=\linewidth]{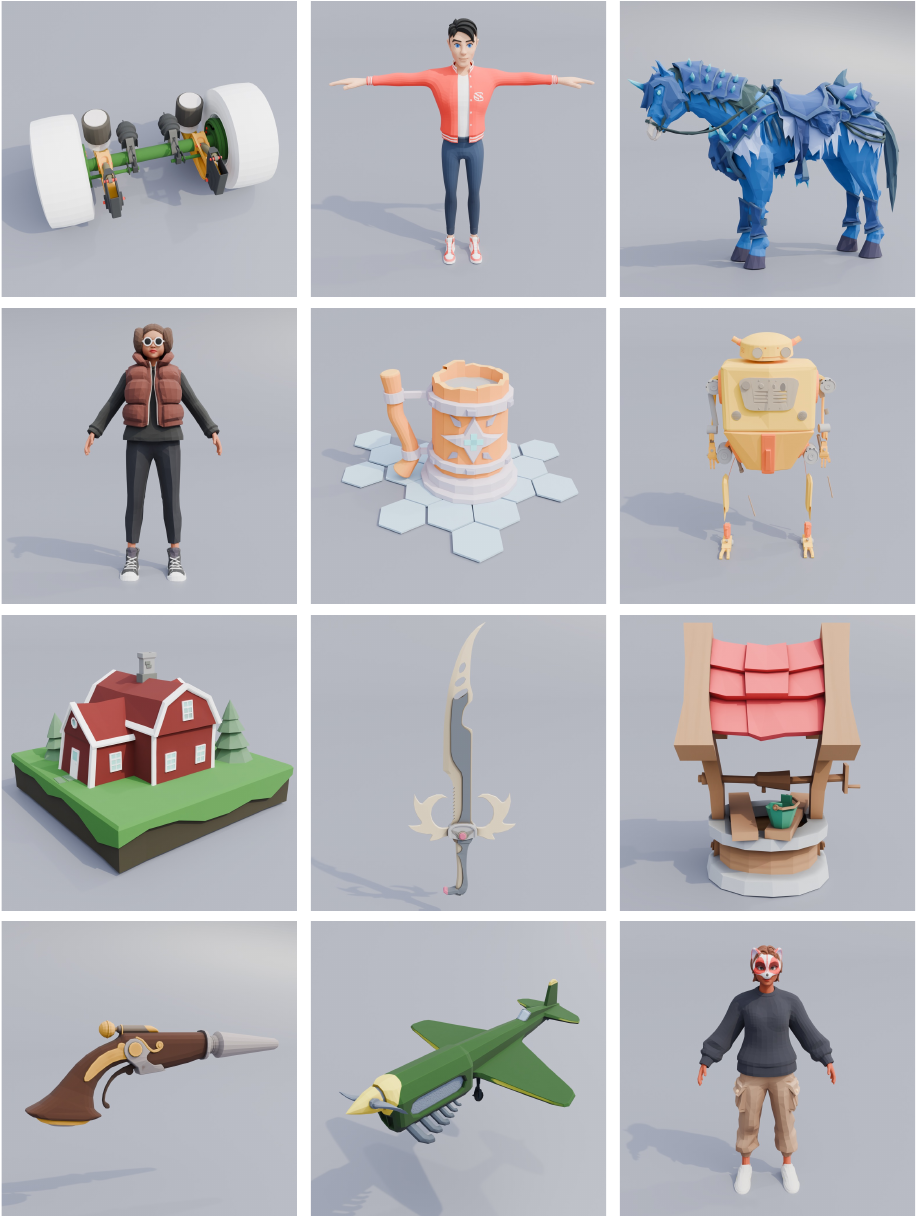}
  \caption{Visual examples of additional part texturing results generated by NaTex.} 
  \label{fig:part_texture_good}
\end{figure}

\textbf{Part Texturing.}
Texturing individual parts is just as straightforward as generating textures for the entire object. Unlike previous MVD approaches, which struggle with interior regions, our method naturally circumvents this issue by predicting color directly in 3D space for different part surfaces. Fig.\ref{fig:part_texture_case} shows part texturing results obtained by directly applying NaTex-2B. It can be observed that our model effectively handles occluded regions between parts and generates accurate textures for these areas. Fig.\ref{fig:part_texture_good} provides additional visual examples.

\begin{figure}[t] 
  \centering
  \includegraphics[width=\linewidth]{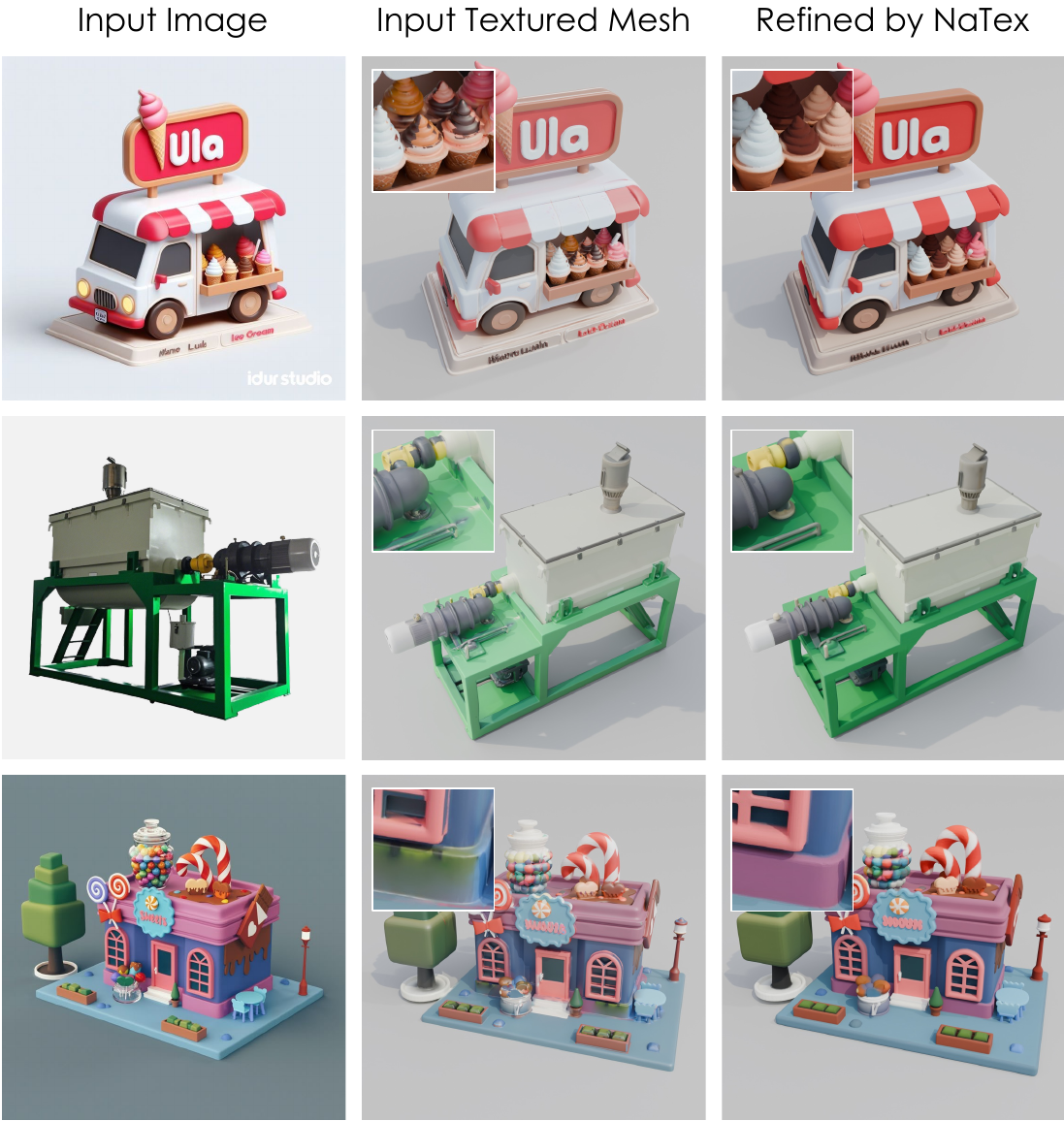}
  \caption{Illustration of texture refinement using NaTex with color control. As shown, NaTex effectively corrects errors in the input mesh caused by occluded regions and inconsistencies.} 
  \label{fig:refinement}
\end{figure}

\textbf{Texture Refinement.}
Our model can also serve as a second-stage refiner for MVD pipelines. This can be easily achieved by fine-tuning NaTex-2B with color control conditioned on an initial texture. In general, our refiner can correct various projection errors and automatically inpaint occluded regions, as illustrated in Fig.~\ref{fig:refinement}. Moreover, thanks to strong conditioning, this process can be performed in just five steps without any distillation, making it extremely fast and efficient for a wide range of downstream tasks.

\section{Limitations and Future Works}

It is exciting that the proposed NaTex advances texture generation, producing more seamless results and generalizing to a variety of applications. However, limitations remain that warrant further research. For example, the reconstruction quality of the VAE could be improved to support higher-resolution textures. Data curation should be enhanced for material generation. Part segmentation could be refined to reduce ambiguity and improve granularity. New methods are needed to handle closed surfaces in adjacent parts for part texturing. Additionally, texture refinement also presents a promising direction for incorporating more 2D priors and leveraging established MVD research.

{
    \small
    \bibliographystyle{ieeenat_fullname}
    \bibliography{sample-base}
}

\end{document}